\documentclass[runningheads]{llncs}
\usepackage{graphicx}
\usepackage{amsmath}
\usepackage{amsfonts}
\usepackage{comment}
\usepackage{placeins}
\usepackage{svg}
\usepackage{hyperref}
\usepackage{floatrow}
\begin{document}
\title{Uncertainty Modelling in Deep Networks: Forecasting  Short and Noisy Series}
\titlerunning{Uncertainty Modelling in DN Forecasting Short and Noisy Series }
%
\author{Axel Brando\inst{1,} \inst{2}, Jose A. Rodr\'iguez-Serrano\inst{1},\\ Mauricio Ciprian\inst{1}, Roberto Maestre\inst{1} and Jordi Vitri\`a\inst{2}}

\authorrunning{Axel Brando et al.}
%
\institute{BBVA Data and Analytics\\
\email{axel.brando@bbvadata.com}, \\\email{joseantonio.rodriguez.serrano@bbvadata.com},
\\\email{mauricio.ciprian@bbvadata.com}\\\email{roberto.maestre@bbvadata.com}
\and
Departament de Matem\`atiques i Inform\`atica \\
Universitat de Barcelona\\
\email{axelbrando@ub.edu}, \email{jordi.vitria@ub.edu}}
\maketitle              
\begin{abstract}
Deep Learning is a consolidated, state-of-the-art Machine Learning tool to fit a function $y=f(x)$ when provided with large data sets of examples $\{(x_i, y_i)\}$.
However, in regression tasks, the straightforward application of Deep Learning models provides a point estimate of the target. In addition, the model does not take into account the uncertainty of a  prediction. This represents a great limitation for tasks where communicating an erroneous prediction carries a risk. In this paper we tackle a real-world problem of forecasting impending financial expenses and incomings of customers, while displaying predictable monetary amounts on a mobile app. In this context, we investigate if we would obtain an advantage by applying Deep Learning models with a Heteroscedastic model of the variance of a network's output. Experimentally, we achieve a higher accuracy than non-trivial baselines. More importantly, we introduce a mechanism to discard low-confidence predictions, which means that they will not be visible to users. This should help enhance the user experience of our product.

\keywords{Deep Learning, Uncertainty, Aleatoric Models, Time-Series}
\end{abstract}

\section{Introduction}
\label{intro}

Digital payments, mobile banking apps, and digital money management tools such as personal financial management apps now have a strong presence in the financial industry. There is an increasing demand for tools  which bring higher interaction efficiency, improved user experience, allow for better browsing using different devices and take or help taking automatic decisions. 

The Machine Learning community has shown early signs of interest in this domain. For instance, several public contests have been introduced since 2015. This notably includes a couple of Kaggle challenges\footnote{https://www.kaggle.com/c/santander-product-recommendation}\footnote{https://www.kaggle.com/c/sberbank-russian-housing-market}\footnote{https://www.kaggle.com/c/bnp-paribas-cardif-claims-management} and the 2016 ECML Data Discovery Challenge. Some studies have addressed Machine Learning for financial products such as recommendation prediction \cite{mitrovic2016predicting}, location prediction \cite{wistuba2016bank} or fraud detection \cite{mutanen2006customer}.

{\bf Industrial context. } Our present research has been carried out within the above-mentioned context of digital money management functions offered via our organization's mobile banking app and web. This includes the latest ``expense and income forecasting'' tool which, by using large amounts of historical data, estimates customers' expected expenses and incomings. On a monthly basis, our algorithm detects recurrent expenses
\footnote{from now on, we will refer to this as 'expenses' because our models treat income as ``negative expenses".} 
(either specific operations or aggregated amounts of expenses within a specific category). We will feed the mobile app with the generated results, which enables customers to anticipate said expenses and, hence, plan for the month ahead.

\begin{figure}
\floatbox[{\capbeside\thisfloatsetup{capbesideposition={right,top},capbesidewidth=4cm}}]{figure}[\FBwidth]
{\caption{Screenshots of BBVA's mobile app showing expected incomes and expenses. Global calendar view (left) and expanded view of one of the forecasts (right).}\label{fig:screen}}
{\includegraphics[width=5cm]{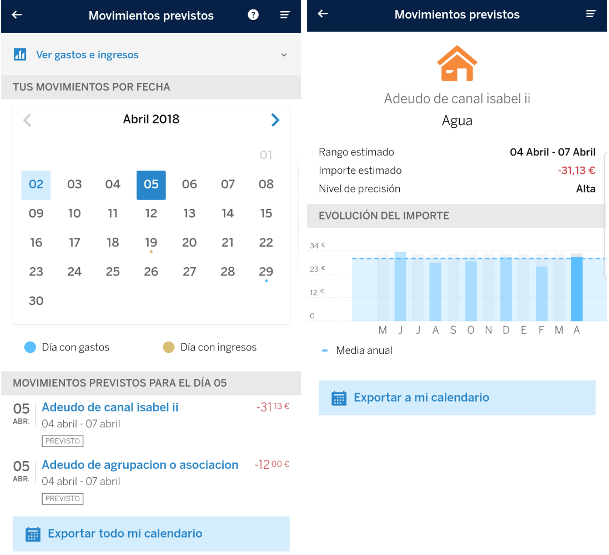}}
\end{figure}

This function is currently in operation on our mobile app and website. It is available  to 5M customers, and generates   hundreds of thousands of  monthly visits. Fig. \ref{fig:screen} displays a screenshot.  

The tool consists of several modules, some of which involve Machine Learning models. A specific problem solved by one of the modules is to estimate the amount of money a customer will spend in a specific financial category in the upcoming month. We have financial category labels attached to each operation. Categories include financial events, as well as everyday products and services such as ATM withdrawals, salary, or grocery shopping, among others (as we will explain hereafter). This problem can be expressed as a regression problem, where the input is a set of attributes of the customer history, while the output is the monetary amount to be anticipated. However, the problem presents several unique challenges. Firstly, as we work with monthly data, the time series are short, and limited to a couple of years; thus, we fail to capture long-term periodicities. Secondly, while personal expenses data can exhibit certain regular patterns, in most cases series can appear erratic or can manifest spurious spikes; indeed, it is natural that not all factors for predicting a future value are captured by past values.

Indeed, preliminary tests with classical time series methods such as the Holt-Winters procedure yielded poor results. One hypothesis is that classical time series method perform inference on a per-series bases and therefore require a long history as discussed in \cite{Hin}.

With the aim of evolving this solution, we asked ourselves whether Deep Learning methods can offer a competitive solution. Deep Learning algorithms have become the state-of-the-art in fields like computer vision or automatic translation due to their capacity to fit very complex functions $\hat{y}=\phi(x)$ to large data sets of pairs $(x,y)$. 

{\bf Our proposed solution.} This article recounts our experience in solving one of the main limitations of Deep Learning methods for regression: they do not take into account the variability in the prediction. In fact, one of the limitations is that the learned function $\phi(x)$ provides a point-wise estimate of the output target, while the model does not predict a probability distribution of the target or a range of possible values. In other words, these algorithms are typically incapable to assess how confident they are concerning their predictions.

This paper tackles a real-world problem of forecasting approaching customer financial expenses in certain categories based on the  historical data available.  
We pose the problem as a regression problem, in which we build features for a user history and fit a function that estimates the most likely expense in the subsequent time period (dependent variable), given a data set of user expenses. Previous studies have shown that Deep Networks provide lower error rates than other models \cite{EUDtimeseries}.

At any rate, Neural Networks are \textit{forced} to take a decision for all the cases. Rather than minimising a forecast error for \textit{all} points, we seek mechanisms to \textit{detect} a small fraction of predictions for each user where the forecast is confident. This is done for two reasons. Since notifying a user about an upcoming expense is a value-added feature, we ask ourselves whether it is viable to reject forecasts for which we are not certain. Furthermore, as a user may have dozens of expense categories, this is a way of selecting relevant impending expenses. 

To tackle the issue of prediction with confidence, potential solutions in the literature combine the good properties of Deep Learning to generically estimate functions with the probabilistic treatments including Bayesian frameworks of Deep Learning (\cite{Kendall}, \cite{BBB}, \cite{MDN}, \cite{DropoutBayesApprox}). Inspired by these solutions, we provide a formulation of Deep regression Networks which outputs the parameters of a certain distribution that corresponds to an estimated target and an input-dependent variance (i.e. Heteroscedastic) of the estimate; we fit such a network by maximum likelihood estimation. We evaluate the network by both its accuracy and by its capability to select ``predictable" cases. For our purpose involving millions of noisy time series, we performed a comprehensive benchmark and observed that it outperforms non-trivial baselines and some of the approaches cited above.

\section{Method}
\label{method}

This section presents the details of the proposed method. 

\begin{figure}[ht]
  \centering
  \includegraphics[width=\textwidth]{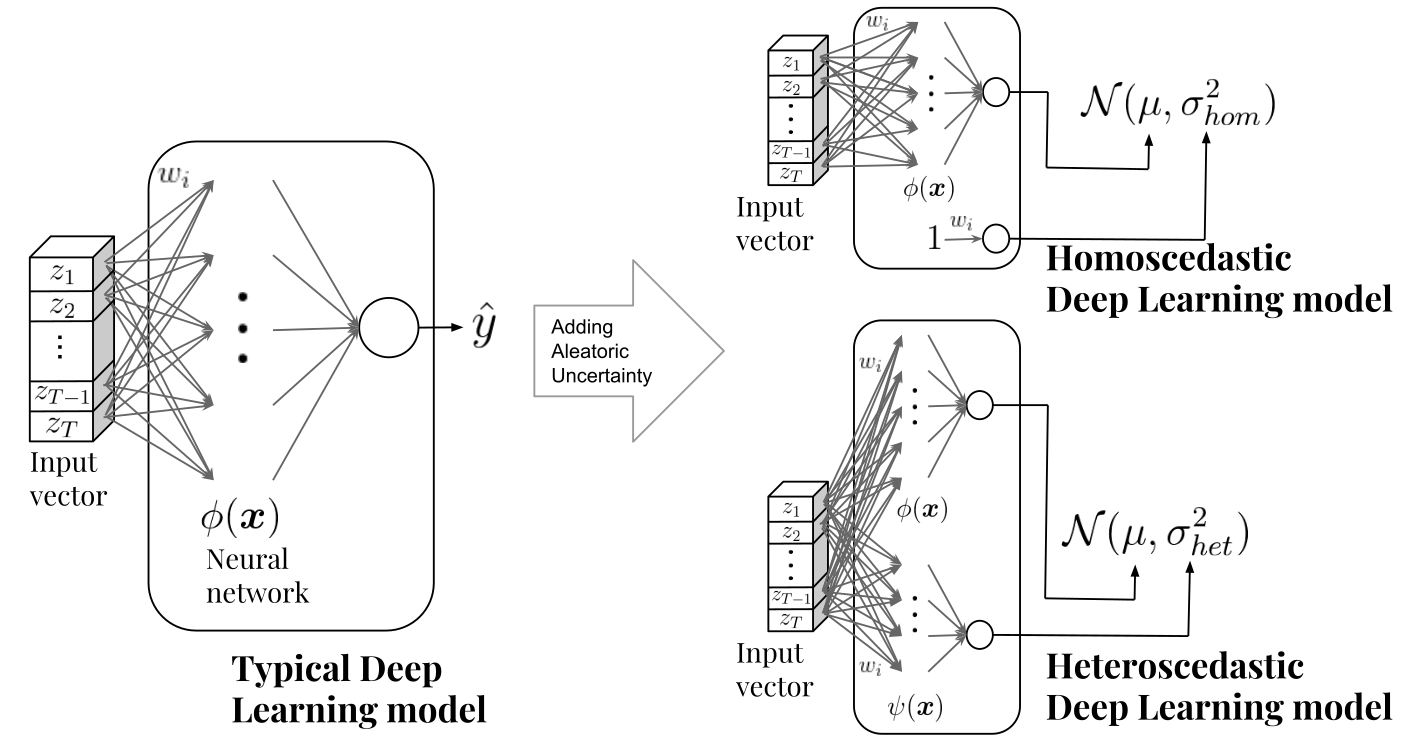}
  \caption{Overview representation of the process to transform a typical Deep Learning model that has a single output given an input vector to a generic Aleatoric Deep Learning model with Gaussian distributions.}
\end{figure}

\subsection{Deep Learning as a point estimation}
\label{subsec:DL}

A generic and practical solution to tackle the regression problem is Deep Learning as it is able to approximate any continuous function following the universal approximation theorem \cite{universal}.

Straightforward implementations of Deep Learning models for regression do not typically consider a confidence interval (or score) of the estimated target.  However, with our particular setting, it is critical that the model identifies the reliable forecasts, and can ignore forecasts where  user-spending gets noisier. Furthermore, if we can find a correlation between the noisy time-series and the error of prediction, a strategy could be devised to improve the general performance of the system. 

In this paper we denote $\hat{y}=\phi_{\boldsymbol{w}}(\boldsymbol{x})$ as the function computed by the Neural Network, with ${\bf x}$ a vector of inputs (the \textit{attributes}) and 
$\hat{y}$ the output, i.e. the forecasted amount. We will denote the set of all weights by $\boldsymbol{w}$. 

In particular, we will use two standard types of layers. On the one hand, we consider non-linear stacking of Dense layers. A Dense layer is a linear combination of the weights $\boldsymbol{w}$ with the input of the layer $\boldsymbol{z}$ passing through an activation function, act, i.e.

$$ \boldsymbol{z}_{n+1} = \text{act} \left( \boldsymbol{w} \boldsymbol{z}_{n} \right) $$

On the other hand, a LSTM layer (\cite{hochreiter1997long}, \cite{gers1999learning}) which, like the other typical recurrent layers \cite{cho2014learning}, maps a time-series with $t$ steps to an output that can have $t$ steps or fewer depending on our needs. In order to do this, for each of the steps, the input $\boldsymbol{z}_n^{(t)}$ will be combined at step $t$ with the same weights and non-linearities for all the steps in the following way:

$$ \boldsymbol{f}_t = sig \left( \boldsymbol{w}_f \cdot [\boldsymbol{h}_{t-1},\boldsymbol{z}_n^{(t)}] + b_f \right)$$
$$ \boldsymbol{i}_t = sig \left( \boldsymbol{w}_i \cdot [\boldsymbol{h}_{t-1}, \boldsymbol{z}_n^{(t)}] + b_i \right) $$
$$ \boldsymbol{c}_t = \boldsymbol{f}_t \ast \boldsymbol{c}_{t-1} + \boldsymbol{i}_t \ast \tanh \left( \boldsymbol{w}_c \cdot [\boldsymbol{h}_{t-1},\boldsymbol{z}_n^{(t)}] + b_c \right) $$
$$ \boldsymbol{o}_t = \text{sig} \left( \boldsymbol{w}_o \cdot [\boldsymbol{h}_{t-1},\boldsymbol{z}_n^{(t)}]^T + b_o \right) $$
$$ \boldsymbol{h}_t = \boldsymbol{o}_t \ast \tanh \left( \boldsymbol{c}_t \right) $$

Where $\boldsymbol{w}_f,b_f,\boldsymbol{w}_i,b_i,\boldsymbol{w}_c, b_c, \boldsymbol{w}_o, b_o$ are the weights of the cell (shared for all the $t$ steps) and sig are Sigmoid functions.


In the experimental section, we consider different Neural Network architectures by stacking Dense and/or LSTM layers. Fitting the network weights ${\bf w}$ involves minimizing a loss function, which is typically based on the standard \textit{back-propagation} procedure. Nowadays, several optimizers are implemented in standard Deep Learning libraries which build on automatic differentiation after specifying the loss. All our Deep Learning models are implemented in Keras \cite{chollet2015keras} with TensorFlow \cite{tensorflow2015-whitepaper} backend and specific loss functions are discussed below.

\subsection{Types of uncertainty}
\label{sec:uncertainty}

 With our system it will be critical  to make highly accurate predictions, and, at the same time, assess the confidence of the predictions. We use the predictions to alert the user to potential impending expenses. Because of this, the cost of making a wrong prediction is much higher than not making any prediction at all. Considering the bias-variance trade-off between the predicted function $\phi$ and the real function to be predicted $f$, the sources of the error could be thought as:

$$ \mathcal{E}(\boldsymbol{x})=\underset{\text{Bias}^{2}}{\underbrace{\left(\mathcal{\mathbb{E}}\left[\phi(\boldsymbol{x})\right]-f(\boldsymbol{x})\right)^{2}}}+\underset{\text{Variance}}{\underbrace{\left(\mathcal{\mathbb{E}}\left[\phi(\boldsymbol{x})-\mathcal{\mathbb{E}}[\phi(\boldsymbol{x})\right]\right)^{2}}}+\underset{\text{cst error}}{\underbrace{\sigma_{cst}}} $$

Where $\mathcal{\mathbb{E}}(z)$ is the expected value of a random variable $z$.

To achieve a mechanism to perform rejection of predictions within the framework of Deep Learning, we will introduce the notion of variance in the prediction. According to the Bayesian viewpoint proposed by \cite{Uncertain}, it is possible to characterise the concept of uncertainty into two categories depending on the origin of the noise. On the one hand, if the noise applies to the model parameters, we will refer to \textit{Epistemic uncertainty} (e.g. Dropout, following \cite{DropoutBayesApprox}, could be seen as a way to capture the Epistemic uncertainty). On the other hand, if the noise occurs directly in the output given the input, we will refer to it as \textit{Aleatoric uncertainty}. Additionally, Aleatoric uncertainty can further be categorised into two more categories: \textit{Homoscedastic uncertainty}, when the noise is constant for all the outputs (thus acting as a ``measurement error"), or \textit{Heteroscedastic uncertainty} when the noise of the output also depends explicitly on the specific input (this kind of uncertainty is useful to model effects such as occlusions / superpositions of factors or  variance in the prediction for an input).

\subsection{A generic Deep Learning regression Network with Aleatoric uncertainty management}

This section describes the proposed framework to improve the accuracy of our forecasting problem by modelling Aleatoric uncertainty in Deep Networks.

The idea is to pose Neural Network learning as a probabilistic function learning. We follow a formulation of Mixture Density Networks models \cite{MDN}, where we do not minimise the typical loss function (e.g. mean squares error), but rather the likelihood of the forecasts.

Following \cite{MDN} or \cite{Kendall}, they define a likelihood function over the output of a Neural Network with a Normal distribution, $\mathcal{N}(\phi(\boldsymbol{x}),\sigma_{ale}^2)$, where $\phi$ is the Neural Network function and $\sigma_{ale}$ is the variance of the Normal distribution. However, there is no restriction that does not allow us to use another distribution function if it is more convenient for our very noisy problem. So we decided to use a Laplace distribution defined as 

\begin{equation}\label{eq:likelihood}
L\mathcal{P}(\boldsymbol{y} \mid \phi(\boldsymbol{x}),b_{ale}) = \frac{1}{2 b_{ale}} \exp \left( - \frac{\mid \boldsymbol{y} - \phi(\boldsymbol{x}) \mid}{b_{ale}}\right),
\end{equation}

which has similar properties and two (location and scale) parameters like the Normal one but this distribution avoids the square difference and square scale denominator of the Normal distribution with an  empirically unstable behaviour in the initial points of the Neural Network weights optimisation or when the absolute error is sizeable. Furthermore, because of the monotonic behaviour of the logarithm, maximising the likelihood is equivalent to minimising the negative logarithm of the likelihood (Eq. \ref{eq:likelihood}), i.e. our loss function, $\mathcal{L}$, to minimise will be as follows

\begin{equation}\label{eq:loss}
\mathcal{L}\left(\boldsymbol{w}, b_{ale}; \{(\boldsymbol{x}_i,y_i)\}_{i=1}^{N} \right) = - \overset{N}{\underset{i=1}{\sum}}\left[-\log(b_{ale})-\frac{1}{b_{ale}}\mid y_{i}-\phi(\boldsymbol{x}_{i})\mid \right]
\end{equation}

where $\boldsymbol{w}$ are the weights of the Neural Network to be optimised.

In line with the above argument, note that 
$b_{ale}$ captures the Aleatoric uncertainty. Therefore, this formulation applies to both the Homoscedastic and Heteroscedastic case. In the former case, $b_{ale}$ is a single parameter to be optimised which  we will denote  $b_{hom}$.

In the latter, the $b_{ale}(\boldsymbol{x})$ is a function that depends on the input. Our assumption is that the features needed to detect the variance behaviours of the output are not directly related with the features to forecast the predicted value. Thus, $b_{ale}(\boldsymbol{x})$ can itself be  the result of another input-dependent Neural Network $b_{ale}(\boldsymbol{x}) = \psi(\boldsymbol{x})$, and all parameters of $\phi$ and $\psi$ are optimised jointly. It is important to highlight that $b_{ale}$ is a strictly positive scale parameter, meaning that we must restrict the output of the $\psi(\boldsymbol{x})$ or the values of the $b_{hom}$ to positive values.

\section{Experimental settings and results}

\subsection{Problem setting}

Our problem is to forecast upcoming expenses in personal financial records with  a data set constructed with
historical account data from the previous 24 months. 
The data was anonymized, e.g. customer IDs are removed from the series, and we do not deal with individual amounts,  but with monthly aggregated amounts. All the experiments were carried out within our servers. 
The data consists of series of monetary expenses for certain customers in  selected expense categories. 

\begin{figure}[ht]
\begin{center}
    \includegraphics[width=\textwidth]{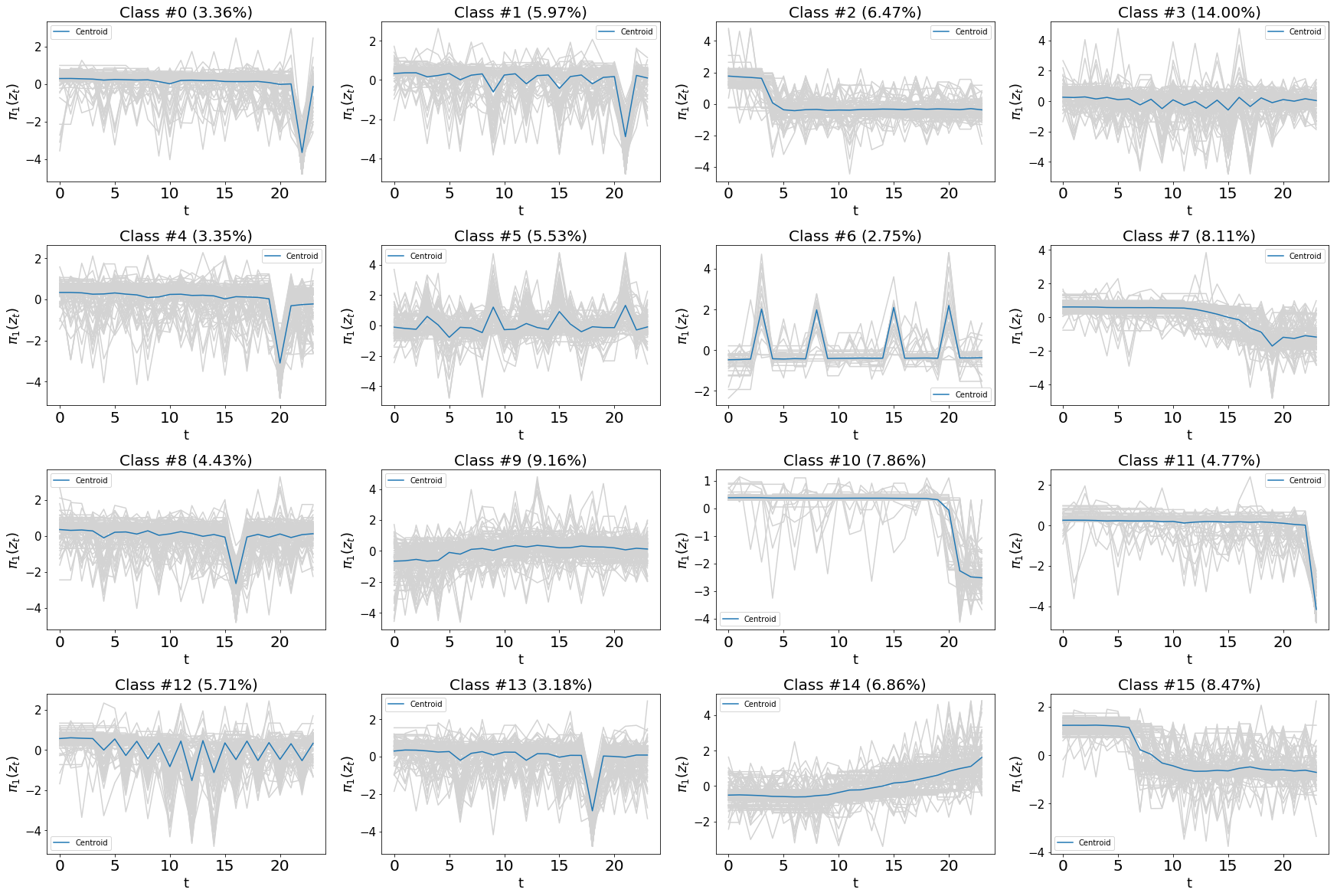}
\end{center}
\caption{Clustering of the normalised $24$ points time-series by using the $\pi_1$ transformation. The grey lines are $100$ samples and the blue line is the centroid for each cluster.}
\label{fig:clustering}
\end{figure}

We cast this problem as a rolling-window regression problem. 
Denoting the observed values of expenses in an individual series over a window of the last $T$ months as ${\boldsymbol{z}}=(z_1,\ldots z_T)$, we extract $L$ attributes $\pi(\boldsymbol{z}) = (\pi_1(\boldsymbol{z}), \ldots, \pi_L(\boldsymbol{z}))$ from ${\boldsymbol{z}}$.   The problem is then to estimate the most likely value of the $(T+1)$th month from the attributes, i.e fit a function for the problem $\hat{y} = \phi(\pi({\boldsymbol{z}}))$, with $y=z_{T+1}$, where we made explicit in the notation that the attributes depend implicitly on the raw inputs. To fit such a model we need a data set of $N$ instances of pairs $(\pi(\boldsymbol{z}), y)$, i.e. $\mathcal{D} = (\pi(\boldsymbol{Z}),\boldsymbol{Y}) = ((\pi(\boldsymbol{z}_1),y_1), \ldots, ({\pi(\boldsymbol{z}_N),y_N}))$.  

To illustrate the nature and the variability of the series, in Fig. \ref{fig:clustering} we visualise the 
values of the raw series 
grouped by the clusters resulting from a $k$-means algorithm where $k=16$. Prior to the clustering, the series were centered by their mean and normalised by their standard deviation so that the emergent clusters indicate scale-invariant behaviours such as periodicity. 

As seen in Fig. \ref{fig:clustering}, the series present clear behaviour patterns. This leads us to believe  that casting the problem as a supervised learning one, using a large data set, is adequate and would capture general recurrent patterns. 

Still, it is apparent that the nature of these data sets presents some challenges, as perceived from the variability of certain clusters in Fig \ref{fig:clustering}, and in some of the broad distributions of the output variable. In many individual cases, an expense may result from erratic human actions, spurious events or depending on factors not captured by the past values of the series. One of our objectives is to have an uncertainty value of the prediction to detect those series - and only communicate forecasts for those cases for which we are confident. 
 
\subsubsection{Data pre-processing details}

 We generate monthly time series by aggregating the data over the contract id + the transaction category label given by an automatic internal  categorizer which employs a set of strategies including text mining classifiers. While this process generates dozens of millions of time series in production, this study uses a random sample of 2 million time series (with $T=24$) for the training set, and 1 million time series for the test set. 
 In order to ensure a head-to-head comparison with the existing system, the random sample is taken only from the subset of series which enter the forecasting module (many other series are discarded by different heuristics during the production process). It is worth noticing that most of these series have non-zero target values. 
From these series we construct the raw data $\boldsymbol{z} = (z_{1}, \ldots, z_{T})$ and the targets $y=z_{T+1}$, from which we compute attributes. 

As attributes, we use the $T$ values of the series, normalized as follows:

$$\pi_1(z_{i})=\begin{cases}
\frac{z_{i}-\overline{\boldsymbol{z}}}{std(\boldsymbol{z})} & \text{if }std(\boldsymbol{z})\geq\theta\\
z_{i}-\overline{\boldsymbol{z}} & \text{if }std(\boldsymbol{z})<\theta
\end{cases}. $$

Where $\theta$ is a threshold value. 

We also add the mean and standard deviation as attributes, which can be written as:  $\boldsymbol{x}_i \equiv (\pi_1(z_{1}),\ldots,\pi_1(z_{T}),\overline{\boldsymbol{z}},std(\boldsymbol{z}))_{i}$. 

The rationale for this choice is: (i) important financial behaviours such as periodicity tend to be scale-invariant, (ii) the mean recovers scale information (which is needed as the forecast is an real monetary value), (iii) the spread of the series could provide information for the uncertainty. We converged to these attributes after a few preliminary experiments. 

\subsection{Evaluation measures}

We evaluate all the methods in terms of an Error vs. Reject characteristic in order to take into account both the accuracy and the ability to reject uncertain samples. Given the point estimates $\hat{y}_i$ and uncertainty scores $\upsilon_i$ of the test set, we can discard all cases with an uncertainty score above a threshold $\kappa$, and compute an error metric only over the set of accepted ones. Sweeping the value of $\kappa$ gives rise to an Error vs. Keep trade-off curve: at any operating point, one reads the fraction of points retained and the resulting error on those -- thus measuring simultaneously the precision of a method and the quality of its uncertainty estimate. 

To evaluate the error of a set of points, we use the well-known mean absolute error (MAE): 

$$\text{MAE}(\boldsymbol{y}, \boldsymbol{x}) = \frac{1}{N} \overset{N}{\underset{i=1}{\sum}} |\boldsymbol{y}_{i}-\phi(\boldsymbol{x}_{i})|$$

which is the mean of absolute deviations between a forecast and its actual observed value. To specify explicitly that this MAE is computed only on the set of samples where the uncertainty score $\upsilon$ is under $\kappa$, we use the following notation: 

\begin{equation}
\label{eq:MAEk}
\text{MAE}_\kappa(\kappa; \boldsymbol{y}, \boldsymbol{x}) = \frac{\overset{N}{\underset{i=1}{\sum}} \left|\boldsymbol{y}_{i}-\phi(\boldsymbol{x}_{i})\right| \cdot \left[ \upsilon_i < \kappa \right] }{\overset{N}{\underset{i=1}{\sum}} \left[\upsilon_i < \kappa \right]}
\end{equation}

And the kept fraction is simply 
${\rm Keep}(\kappa) = {\sum} \left[\upsilon_i < \theta \right]/N$. Note that $\text{MAE}_\kappa = \text{MAE}$ when $\kappa$ is the maximum $\upsilon$ value.

Our goal in this curve is to select a threshold such that the maximum error is controlled; since large prediction errors would affect the user experience of the product. In other words, we prefer not to predict everything  than to predict erroneously.

Hence, with the best model we will observe the behaviour of the MAE error and order the points to predict according to the uncertainty score $\upsilon(\boldsymbol{x})$ for that given model.

In addition, we will add visualisations to reinforce the arguments of the advantages of  using the models described in this paper that take into account Aleatoric uncertainty to tackle noisy problems.

\subsection{Baselines and methods under evaluation}

We provide details of the methods we evaluate as well as the baselines we compare are comparing against. {\bf Every method below outputs an estimate of the target along with an uncertainty score}. We indicate in italics how each estimator and uncertainty score is indicated in Table \ref{tab:ErrorReject}. For methods which do not explicitly compute uncertainty scores, we will use the variance $\text{var}(\boldsymbol{z})$ of the input series as a proxy. 

\vspace{0.15cm}
{\bf Trivial baselines. }
In order to validate that the problem  cannot be easily solved with simple forecasting heuristics such as a moving average, we first evaluate three simple predictors: (i) the mean of the series $\hat{y}=\overline{\boldsymbol{z}}$ ({\it mean}), (ii) $\hat{y}=0$ ({\it zero}), and (ii) the value of the previous month $\hat{y}=z_{T}$ ({\it last}). In the three cases, we use the variance of the input as the uncertainty score $\text{var}(\boldsymbol{z})$ ({\it var}).   

\vspace{0.15cm}
{\bf Random Forest. }
We also compare against the forecasting method which is currently implemented in the expense forecasting tool. The tool consists of a combination of modules. The forecasting module is based on random forest ({\it RF*}). Attributes include the values of the 24 months ($z$) but also  dozens of other carefully hand-designed attributes. The module outputs the forecasted amount and a discrete confidence label (low/medium/high), referred to as {\it prec}. For the sake of completeness, we also repeat the experiment using the variance of the input ({\it var}) as the confidence score. 

\vspace{0.15cm}
{\bf General Additive Model. }
We also compare against a traditional regression method able to output estimates and their distribution, specifically we use the Generalized Additive Models ({\it GAM}) \cite{AndersonGAM} with its uncertainty score denoted as {\it SE}.  {\it GAM} is a generalized lineal model with a lineal predictor involving a sum of smooth functions of covariates \cite{HastTibs1990} that follows the following form,

$$
g(\mu_i)  = X^{*}_{i} \theta + f_1(x_{1i}) + f_1(x_{2i}) + f_3(x_{3i},x_{4i}) + ...$$

where $\mu_i \equiv \mathbb{E}(Y_i)$, $Y_i$ is the response variable that follows some exponential family distribution, $X^{*}_{i}$ is a row of the model matrix for any strictly parametric model components, $\theta$ is the parametric vector and $f_{j}$ are smooth functions of the covariates, $x_j$, and $g$ is a known link function. We fit {\it GAM} in a simple way: $g(\mu_i)  = f_1(\boldsymbol{z}_1) + ... + f_N(\boldsymbol{z}_N)$. The parameters in the model are estimated by penalized iteratively re-weighted least squares (P-IRLS) using Generalized Cross Validation (GCV) to control over-fitting.

\vspace{0.15cm}
{\bf Deep Learning Models. }
Finally, we will compare a number of Deep Learning models with various strategies to obtain the forecasted amount and an uncertainty score.

\vspace{0.15cm}
{\it Dense networks. } We start with a plain Dense Network architecture (referred to as \textit{Dense}). In particular, our final selected Dense model is composed by 1 layer of 128 neurons followed by another of 64 neurons, each of the layers with a ReLu activation, and finally this goes to a single output. We use the Mean Absolute Error (MAE) as loss function. Since a Dense regression Network does not provide a principled uncertainty estimate, again we use the variance of the series, $\text{var}({\bf z})$, as the uncertainty score. 

\vspace{0.15cm}
{\it Epistemic Models.} We also evaluate Bayesian Deep Learning alternatives, which offer a principled estimate of the uncertainty score. These carry out an {\it Epistemic} treatment of the uncertainty (see Section \ref{sec:uncertainty}), rather than our \textit{Aleatoric} treatment.  We consider two methods: (i)  adding Dropout \cite{DropoutBayesApprox} (with a 0.5 probability parameter of Dropout)  (referred to as \textit{DenseDrop}), and (ii) Bayes-By-Backprop variation \cite{BBB}, denoted as \textit{DenseBBB}. Both techniques model the distribution of the network weights, rather than weight values. This allows us to take samples from the network weights. Therefore, given a new input, we obtain a set of output samples rather than a single output; we can take the mean of those samples as the final target estimate and its standard deviation as our uncertainty score, i.e. $\upsilon(\boldsymbol{x}) = std \left( \{\phi_{w_i}(\boldsymbol{x})\}_i \right)$. In Table \ref{tab:ErrorReject}, we denote the uncertainty score of Dropout and BBB as \textit{drop} and \textit{BBB}, respectively. For the sake of completeness, we also consider the simple case of the variance (\textit{var}). 

\vspace{0.15cm}
{\it Homoscedastic and Heteroscedastic Models.}
Here we consider fitting a network under the Aleatoric model (see \ref{eq:loss}), both for 
the Homoscedastic and Heteroscedastic 
cases. It is important to highlight that, in both cases, the way used to restrict the values of $b_{ale}$ to positive values was by applying a function $g(b_{ale})$ to the values of the $b_{hom}$ or output of $\psi(\boldsymbol{x})$ defined as the translation of the ELU function plus 1,

$$g(x)=ELU(\alpha,x)+1=\begin{cases}
\alpha(e^{x}-1)+1 & \mbox{for }x<0\\
x+1 & \mbox{for }x\geq0
\end{cases}$$

In addition, in the Heteroscedastic case, we used the same Neural Network architecture for the ``variance" approximation $\psi(\boldsymbol{x})$ than for the ``mean" approximation $\phi(\boldsymbol{x})$. 

In the Homoscedastic case, since the variance is constant for all the samples, we have to resort to the variance of the series as the uncertainty scores. In contrast, the Heteroscedastic variance estimate is input-dependent and can thus be used as the uncertainty score (denoted as \textit{$b_{het}$}) in Table \ref{tab:ErrorReject}. 

All the Deep Learning models were trained during $800$ epochs with Early Stopping by using the $10\%$ of the training set as a validation set. 

\vspace{0.15cm}
{\it LSTM networks.}
To conclude, we will repeat the experiment by replacing the Dense Networks with long short-term memory (\textit{LSTM}) networks, which are more suited to sequential tasks. Here we consider both types of aletoric  uncertainty (i.e. \textit{LSTMHom} and \textit{LSTMHet}). In this particular case, the architecture uses two LSTM layers of 128 neurons followed by a 128 neurons Dense layer, and finally a single output. 

\subsection{Results}

 \begin{figure}[t]
\begin{center}
    \includegraphics[width=1.0\textwidth]{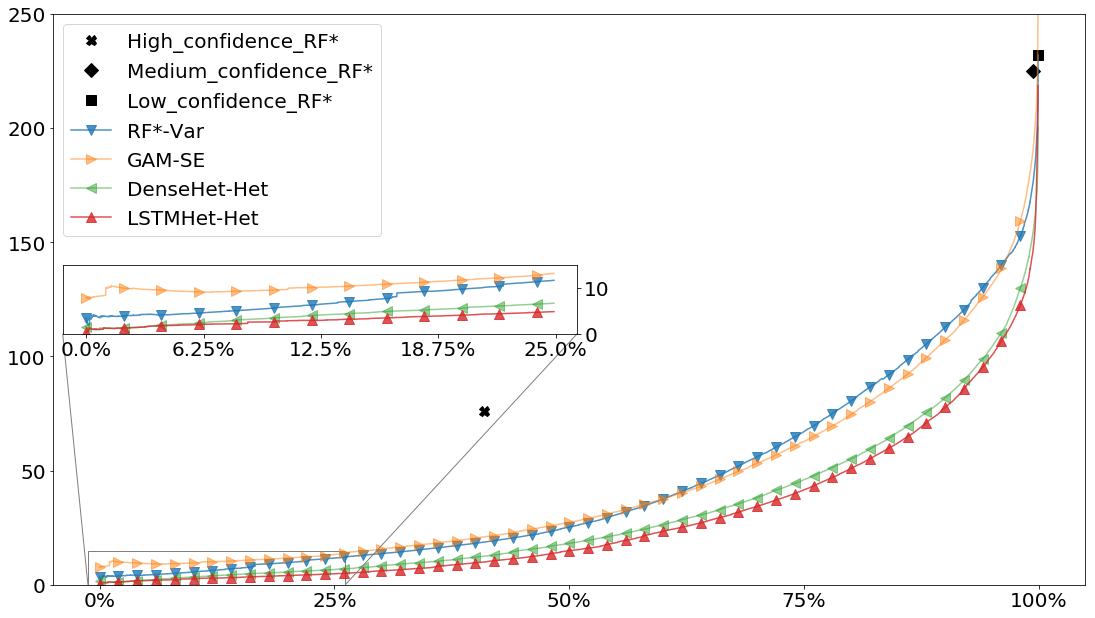}
\end{center}
\caption{Error-keep curve: MAE versus fraction of the samples kept obtained by selected forecasting methods, when cutting at different thresholds of their uncertainty scores. }
\label{fig:ER}
\end{figure}

{\bf Models comparison.}  In Figure \ref{fig:ER} we present a visual comparison between the Error-Keep curves (Eq. \ref{eq:MAEk}) of RF* (indicating the points of low/medium/high confidence),  RF* using \textit{var} as confidence score, the GAM (which uses its own uncertainty estimate), and the best Dense and LSTM models (which, as will be detailed later, use the Heteroscedastic uncertainty estimates).

First we notice that the Heteroscedastic solutions are better than all the other previous solutions. This confirms our assumption that, for our given noisy problem, taking into account the variability of the output provides better accuracy at given retention rates.

\begin {table}[ht]
\caption {Errors (MAE) of each method at points of the Error-Keep curve corresponding to Keep=K. Each row corresponds to the combination of a predictor and an uncertainty score (described in the main text). } 
\label{tab:ErrorReject} 
\begin{center}
\begin{tabular}{ |p{3cm}||p{1.3cm}|p{1.3cm}|p{1.3cm}|p{1.3cm}|p{1.3cm}|p{1.3cm}|p{1.3cm}|  }
 \hline
 {\tiny Predictor + uncertainty}& K=25\% & K=41\% & K=50\% & K=75\% & K=99.5\% & K=100\%\\
 \hline
 $mean$ + $var$   & $36.93$ & $54.09$& $67.54$ & $152.84$ & $417.93$ & $491.68$\\
 Zeros + $var$ & $36.95$ & $54.21$& $67.75$ & $154.16$ & $429.29$ & $504.45$\\
 Last + $var$ & $50.25$ & $75.19$& $94.53$ & $203.27$ & $474.12$ & $606.58$\\
 GAM + $var$ & $16.37$ & $23.41$& $29.66$ & $68.83$ & $191.73$ & $7539.91$\\
 GAM + $SE$ & $13.19$ & $20.97$& $27.30$ & $62.62$ & $191.86$ & $7539.91$\\
 RF* + prec & N/A & $76.08$& N/A & N/A & $224.95$ & $232.11$\\
 RF* + $var$ & $11.69$ & $18.68$& $25.21$ & $66.90$ & $177.15$ & $232.11$\\
 \hline
 Dense + $var$ & $12.3\pm .24$ & $18.1\pm .26$& $23.5\pm .26$ & $56.3\pm .28$ & $146.\pm .48$ & $193.\pm .49$\\
 DenseDrop + $var$ & $14.9\pm .19$ & $20.8\pm .13$& $26.4\pm .07$ & $61.1\pm .30$ & $160.\pm .80$ & $208.\pm 1.0$\\
 DenseDrop + Drop & $16.1\pm 1.3$ & $23.2\pm .91$& $29.5\pm .76$ & $64.9\pm .62$ & $162.\pm 1.4$ & $208. \pm 1.0$\\
 DenseBBB + $var$ & $16.1\pm 1.6$ & $23.5\pm 1.9$& $30.6\pm 2.2$ & $79.8\pm 4.4$ & $198.\pm 5.2$ & $248.\pm 5.6$\\
 DenseBBB + BBB & $19.9\pm 1.1$ & $28.5\pm 1.2$& $36.0\pm 1.7$ & $89.4\pm 4.6$ & $199.\pm 4.7$ & $248.\pm 5.6$\\
 DenseHom + $var$ & $12.3\pm .23$ & $18.1\pm .22$& $23.5\pm .22$ & $56.3\pm .30$ & $145.\pm .89$ & $193.\pm .84$\\
 DenseHet + $var$ & $10.7\pm .08$ & $16.4\pm .10$& $21.7\pm .11$ & $55.0\pm .30$ & $150.\pm .97$ & $199.\pm .84$\\
 DenseHet + $b_{het}$ & $8.24\pm 1.3$ & $12.9\pm .90$& $18.1\pm .59$ & $45.3\pm .67$ & $153.\pm 1.1$ & $199.\pm .84$\\
 \hline
 LSTM + $var$ & $11.6\pm .13$ & $17.6\pm .12$& $23.1\pm .13$ & $55.7\pm .43$ & $146.\pm .81$ & $205.\pm 1.2$\\
 LSTMHom + $var$ & $11.9\pm .37$ & $18.0\pm .59$& $23.5\pm .76$ & $56.6\pm 2.0$ & $150. \pm 9.1$ & $210. \pm 14.$\\
 LSTMHet + $var$ & $10.5\pm .03$ & $16.2\pm .07$& $21.6\pm .10$ & $53.9\pm .17$ & $147.\pm 1.3$ & $218.\pm 1.6$\\
 LSTMHet + $b_{het}$& $5.04\pm .23$ & $10.7 \pm .41$& $15.2 \pm .24$ & $41.1 \pm .54$ & $149. \pm 2.9$ & $218. \pm 1.6$\\
 \hline
\end{tabular}
\end{center}
\end {table}

For a more detailed analysis, Table \ref{tab:ErrorReject} shows the MAE values of all the compared methods for several cut-off points of the Error-Keep curve, namely for keep=25\%,50\%,75\%,100\%, and additionally for 41\% and 99.5\% (for direct comparison to the currently implemented RF* method, because these are the cut-off values obtained by the RF* method with confidence=high and confidence=medium). \textit{Note that we are more interested in low values of keep (typically below 50\%)}, as they correspond to ``selecting'' the most confident samples; but we show all the percentages for completeness.

For  Deep Learning models, to avoid a known issue of sensitivity with respect to the random seeds,  we repeated all experiments $6$ times  with different random  initialisations and report the mean and standard deviation over the 6 runs.

Table \ref{tab:ErrorReject} shows different interesting aspects. First, we confirm that the best performing model is the LSTM network with the proposed Heteroscedastic uncertainty treatment (except in the points where keep $>$ 99.5\%. However, these are not of practical importance because there is virtually no rejection, and the total number of errors are high for all the methods). Had we not considered LSTM networks as an option, the best performing model would still be a Dense Network with Heteroscedastic uncertainty. We also observed that the performance ranking of the incremental experiments between Dense and LSTM is consistent. 

We also note that, for this particular task, the Aleatoric methods (both Homoscedastic and Heteroscedastic) perform better than Epistemic models such as Dropout or BBB. We believe this to be specific to the task at hand where we have millions of short and noisy time series. We conclude that the variability of human spending, with complex patterns, erratic behaviors and intermittent spendings, is captured more accurately by directly modelling an input-dependent variance with a complex function, than by considering model invariance (which may be better suited for cases where the data is more scarce or the noise smoother). 

Another interesting observation is that the  Random Forest baseline used more attributes than the proposed solutions based on Deep Learning, which just used $T$ values of the historical time series. This confirms that fitting a Deep Network on a large data set is a preferred solution; even more when being able to model the uncertainty of the data, as is the case here.

Last but not least, comparing the same uncertainty score with a certain model and its Heteroscedastic version we also observe that there is an accuracy improvement. This means that taking into account uncertainty in the training process, in our noisy problem, not only gives to us an uncertainty score to reject uncertain samples but also it helps in order to predict better.

\begin{figure}[t]
\begin{center}
    \includegraphics[width=0.7\textwidth]{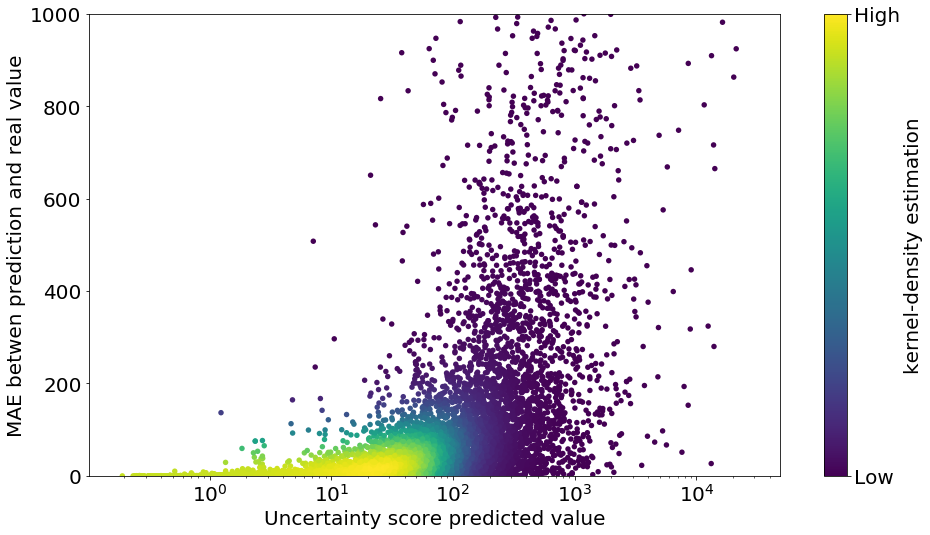}
\end{center}
\caption{Correlation between MAE of prediction and the real $T+1$ value and its uncertainty score of random selection of $10,000$ time series of the test set. The colours represent the density of the zones by using Gaussian kernels.}
\label{fig:CF}
\end{figure}
{\bf Error-Uncertainty score correlation. }
An interesting question is whether or not there exists any relationship between the errors of a model and the uncertainty  scores $\upsilon$ it provides. In Figure \ref{fig:CF} we show the correlation between the MAE of $10.000$ randomly selected points of the test set and the uncertainty score of the best Heteroscedastic model expressed in the logarithm scale. To reduce the effect of clutter, we used a colour-map for each point that represents the density of the different parts of the Figure as resulting from a Gaussian Kernel Density Estimate.  

This figure exhibits two main regimes. On the one hand, for high uncertainty scores ($> 100$), we observe scattered samples (considering their purple colour) where the errors range from very low to very high values and do not seem to align with the uncertainly scores. Upon inspection, many of these series correspond to what humans would consider `unpredictable'': e.g. an expense in month $T+1$ which was much larger than any of the observed $T$ expenses, or series without a clear pattern. On the other hand, we observe a prominent high-concentration area (the yellow-ish one) of low-error and low-uncertainty values; indicating that, by setting the threshold under a specific value of the  uncertainty score, the system mostly selects low-error samples. 

\section{Related Work}

In the case of using Deep Learning for classification, it is common to use a Softmax activation function in the last layer \cite{softmax}. This yields a non-calibrated probability score  which can be used heuristically as the confidence score or calibrated to a true probability \cite{Hendrycks}. However, in the case of regression, the output variables are not class labels and it is not possible to obtain such scores. On the other hand, \cite{Kendall} introduces the idea of combining different kinds of Aleatoric and Epistemic uncertainties. Nevertheless, as we saw in Table \ref{tab:ErrorReject}, the use of Epistemic uncertainty for our problem worsens the accuracy and the general performance of the model.

While there have been several proposals to deal with uncertainties in Deep Learning, they boil down to two main families. On the one hand, some approaches consider the uncertainty of the output. Typically, they construct a network architecture so that its output is not a point estimate, but a distribution \cite{MDN}. On the other hand, some approaches consider the uncertainty of the model. These apply a Bayesian treatment to the optimisation of the parameters of a network (\cite{DropoutBayesApprox}, \cite{BBB}, \cite{Hern}, \cite{MCofBL}). In the present work we observe that if a highly noisy problem can change the loss function in order to minimise a likelihood function as in \cite{MDN}, and by introducing some changes explained above, we are provided with a significant improvement which is crucial to identifying forecasts with a high degree of confidence and even improve the accuracy.

We are dealing with a problem that contains high levels of noise. To make predictions is therefore risky. These are the two reasons why it was our goal to find a solution that improves the mean-variance solution that can be a ``challenging" solution. In order to do that, we grouped several theories proposed in \cite{Kendall} and \cite{MDN} to create a Deep Learning model that takes into account the uncertainty and provides a better performance.

\section{Conclusion}

We explore a new solution for an industrial problem of forecasting real expenses of customers. Our solution is based on Deep Learning models for effectiveness and solve the challenge of uncertainty estimation by learning both a target output and its variance, and 
performing maximum likelihood estimation of the resulting model that contains one network for the target output and another for its variance. We show that this solution obtains better error-reject characteristics than other (traditional and Deep) principled models for regression uncertainty estimation, and outperforms the characteristic that would be obtained by the current industrial system in place. 
While Epistemic models such as Dropout or BBB did not improve the performance in this specific task, we are already working in combining them with our Aleatoric treatment to consider both types of uncertainty in the same model. This is considered future work. We also highlight that, while the present model seems to be able to detect confident predictions, it still lacks mechanisms to deal with the ``unknown unknowns'' problem; and believe that incorporating ideas such as those in \cite{Hendrycks} may help in future work. 

\paragraph{Acknowledgements}
We gratefully acknowledge the Industrial Doctorates Plan of Generalitat de Catalunya for funding part of this research. The UB acknowledges the support of NVIDIA Corporation with the donation of a Titan X Pascal GPU and recognizes that part of the research described in this chapter was partially funded by TIN2015-66951-C2, SGR 1219. We also thank Alberto R\' ubio and C\'esar de Pablo for insightful comments as well as BBVA Data and Analytics for sponsoring the industrial PhD. 

%
%
%
\bibliographystyle{splncs04}
\bibliography{bibliography}
\end{document}